\documentclass[11pt]{article}
\usepackage{style/naaclhlt2016}
\usepackage{times}
\usepackage{latexsym}
\usepackage{amsmath}
\usepackage{graphicx}
\usepackage{color}
\usepackage{xspace}
\usepackage{array}
\makeatletter
\newcommand{\@BIBLABEL}{\@emptybiblabel}
\newcommand{\@emptybiblabel}[1]{
\makeatother}
\newcolumntype{P}[1]{>{\centering\arraybackslash}p{#1}}
\usepackage{hyperref}
\usepackage{url}

\usepackage[aboveskip=-2pt,belowskip=+3pt]{subcaption}
\naaclfinalcopy

\author{
	Justine Zhang,$^{1}$ Ravi Kumar,$^{2}$ Sujith Ravi,$^{2}$ Cristian Danescu-Niculescu-Mizil$^{1}$\\
        $^1$Cornell University, $^2$Google\\
  	\small {\tt \href{mailto:jz727@cornell.edu}{jz727@cornell.edu}},
	{\tt \href{mailto:ravi.k53@gmail.com}{ravi.k53@gmail.com}}, \\
	\small {\tt \href{mailto:ravi.sujith@gmail.com}{ravi.sujith@gmail.com}},
	{\tt \href{mailto:cristian@cs.cornell.edu}{cristian@cs.cornell.edu}} \\
}

\newcommand{\cut}[1]{}
\definecolor{wine}{rgb}{0.65,0,0.35}

\newcommand{\cd}[1]{{\textcolor{blue}{}}}
\newcommand{\justine}[1]{{\textcolor{wine}{}}}

\newcommand{\xhdr}[1]{{\noindent\bfseries #1.}}
\newcommand{\talkingpoints}{talking points\xspace}
\newcommand{\talkingpoint}{talking point\xspace}
\newcommand{\Talkingpoints}{Talking points\xspace}
\newcommand{\fw}[1]{\ensuremath{\mathcal{W}_{#1}}}
\newcommand{\freqtp}[3]{\ensuremath{f_{#1}(#2,#3)}}
\newcommand{\discusspoints}{discussion points\xspace}
\newcommand{\discusspoint}{discussion point\xspace}

\newcommand{\selfcoverage}{self-coverage\xspace}
\newcommand{\opponentcoverage}{opponent-coverage\xspace}
\newcommand{\Intro}{\mathrm{Intro}}
\newcommand{\Disc}{\mathrm{Disc}}

\title{Conversational Flow in Oxford-style Debates}

\begin{document}

\maketitle
\begin{abstract}

Public debates are a common platform for presenting and juxtaposing diverging views on important issues. In this work we propose a methodology for tracking how ideas flow between participants throughout a debate. We use this approach in a case study of Oxford-style debates---a competitive format where the winner is determined by audience votes---and show how the outcome of a debate depends on aspects of conversational flow.
In particular, we find that winners tend to make better use of a debate's interactive component than losers, 
by actively pursuing their opponents' points rather than promoting their own ideas over the course of the conversation.

\end{abstract}

\section{Introduction}
\label{sec:intro}
Public debates are a common platform for presenting and juxtaposing diverging viewpoints
As opposed to 
monologues where 
speakers
are limited to 
expressing their own 
beliefs, debates allow for participants to interactively attack their opponents' points while defending their own. The resulting flow of ideas is a key feature of this conversation genre.

In this work we introduce a computational framework for characterizing debates in terms of conversational flow.
This framework captures two main debating strategies---promoting one's own points and attacking the opponents' points---and tracks 
their relative usage
throughout the debate. By applying this 
methodology
 to a 
 setting where debate winners are known, we show that 
conversational flow patterns are predictive of 
which debater is more likely to persuade an audience.

\xhdr{Case study: Oxford-style debates}
Oxford-style debates provide a setting that is particularly convenient for studying the effects of conversational flow. In this 
competitive debate format, two teams argue for or against a preset motion in order to persuade a live audience to take their position. The audience votes before and after the debate, and the winning team is the one that sways more of the audience towards its view. This 
setup allows us to focus on the effects of conversational flow since it disentangles them from the audience's prior leaning.%
\footnote{Other potential confounding factors are mitigated by the tight format and topic enforced by the debate's moderator.}

The debate format involves an opening statement from the two sides, which presents an overview of their arguments before 
the discussion begins.
 This allows us to 
easily identify talking points held by the participants prior to the interaction, and %
consider them separately from points introduced spontaneously to serve the discussion.

This work is taking steps towards better modeling of conversational dynamics, by: 
(i) introducing a debate dataset with rich metadata (Section~\ref{sec:data}), 
(ii) proposing a framework for tracking the flow of ideas (Section~\ref{sec:method}), and
(iii) showing its effectiveness in a predictive setting (Section~\ref{sec:eval}).

\section{Debate Dataset: Intelligence Squared}
\label{sec:data}
In this study we use transcripts and results of Oxford-style debates from the public debate series ``Intelligence Squared Debates'' (IQ2 for short).\footnote{\scriptsize\url{http://www.intelligencesquaredus.org}}  
These debates are recorded live, and contain motions covering a diversity of topics ranging from foreign policy issues to the benefits of organic food.  Each debate consists of two opposing teams---one for the motion and one against---of two or three experts in the topic of the particular motion, along with a moderator.  Each debate follows the Oxford-style format and consists of three rounds. In the \emph{introduction}, each debater is given 7 minutes to lay out their main points.  During the \emph{discussion}, debaters take questions from the moderator and audience, and respond to attacks from the other team. This round lasts around 30 minutes 
and is highly interactive; teams frequently engage in direct conversation with each other.  Finally, in the \emph{conclusion}, each debater is given 2 minutes to make final remarks.

Our dataset consists of the transcripts of all debates held by IQ2 in the US from September 2006 up to September 2015; in total, there are 108 debates.\footnote{We omitted one debate due to pdf parsing errors.}   
Each debate is quite extensive: on average, 12801 words are uttered in 117 turns by members of either side per debate.\footnote{The processed data is available at \scriptsize\url{http://www.cs.cornell.edu/~cristian/debates/}.}  

\xhdr{Winning side labels}
We follow IQ2's criteria for deciding who wins a debate, as follows. Before the debate, the live audience votes on whether they are for, against, or undecided on the motion. A second round of voting occurs after the debate. A side wins the debate if the difference between the percentage of votes they receive 
post- and pre-debate
 (the ``{delta}'') is greater than that of the other side's.  Often the debates are quite tight: for 30\% of the debates, the difference between the winning and losing sides' deltas is less than 10\%.

\xhdr{Audience feedback}
We check that the voting results are meaningful by verifying that audience reactions to the debaters are related to debate outcome. Using laughter and applause received by each side in each round\footnote{Laughter and applause are indicated in the transcripts.} as markers of positive reactions, we note that differences in audience reception of the two sides emerge over the course of the debate. 
While both sides get similar levels of reaction during the introduction, winning teams tend to receive more laughter during the discussion ($p<0.001$)\footnote{Unless otherwise indicated, all reported $p$-values are calculated using the Wilcoxon signed-rank test.} and more applause during the conclusion ($p=0.05$).

\xhdr{Example debate}
We will use a debate over the motion ``Millennials don't stand a chance'' (henceforth {\em Millennials}) as a running example.\footnote{\scriptsize\url{http://www.intelligencesquaredus.org/debates/past-debates/item/1019-millennials-dont-stand-a-chance}}  
The For side won the debate with a delta of 20\% of the votes, compared to the Against side which only gained 5\%.

\section{Modeling Idea Flow}
\label{sec:method}
Promoting one's own points and addressing the opponent's points are two primary debating strategies.  Here we introduce a methodology to identify these strategies, and use it to investigate their usage and effect on a debate's outcome.\footnote{In the subsequent discussion, we treat all utterances of a particular side as coming from a single speaker and defer modeling interactions within teams to future work.}  %
\begin{table*}[t]
\begin{center}
\begin{tabular}{|l | p{6.75cm} | p{6.75cm} |}
\hline
\footnotesize
Talking point & \multicolumn{1}{c|}{\footnotesize volunteer} & \multicolumn{1}{c|}{\footnotesize boomer}\\
\hline
\footnotesize Introduction & \footnotesize AGAINST: [millennials] \textbf{volunteer}
more than any generation. 73 percent of millennials \textbf{volunteered} for a nonprofit in
2012.  And the percentage of [students] believing that it's [...]
important to help people in need is [at the highest level] in 40 years. & \footnotesize FOR: [\textit{referring to college completion rate}] the \textbf{boomer} generation is now [at] 32 percent. [Millennials] are currently at [...] 33 percent. So this notion that [millennials] have more education at this point in time than anybody else is not actually true.\\
\footnotesize
Discussion & \footnotesize FOR: I'd make the argument [that] \textbf{volunteering} [is done] for exntrinsic [sic] reasons. So, it's done for college applications, or it's done because it's a requirement in high school. & \footnotesize FOR: It stinks to be young, having gone through what your generation [\textit{referring to millennials}] has gone through. But keep in mind that  [...] have gone through the same. \\
\hline
\end{tabular}
\end{center}
\caption{Example \talkingpoints{} used throughout the ``Millennials'' debate. Each talking point belongs to the side uttering the first excerpt, taken from the introduction; the second excerpt is from the discussion section. In the first example, the For side addresses the opposing \talkingpoint{} \textbf{volunteer} during the discussion; in the second example the For side refers to their own \talkingpoint{} \textbf{boomer} and recalls it later in the discussion.\label{tab:tp_example}}
\end{table*}

\xhdr{Identifying \talkingpoints{}}
We first focus on ideas which form the basis of a side's stance on the motion. 
We identify such \emph{\talkingpoints{}} by considering words whose frequency of usage differs significantly between the two teams 
during the introduction, before any interaction takes place. 
To find these words, we use the method introduced by \newcite{monroe2008fightin} in the context of U.S. Senate speeches. In particular, we estimate the divergence between the two sides' word-usage in the introduction, where word-usage is modeled as multinomial distributions smoothed with a uniform Dirichlet prior,
 and divergence is given by 
  log-odds ratio. The most discriminating words are 
those with the highest and lowest z-scores of divergence estimates. 
For a side $X$, we define the set of \talkingpoints{} $\fw{X}$ to be the $k$ words with the highest or lowest $z$-scores.\footnote{In order to focus on concepts central to the sides' arguments, we discard stopwords, perform stemming on the text, and take $k=20$. We set these parameters by examining one subsequently discarded debate.}
 We distinguish between $X$'s {\em own} \talkingpoints{} $\fw{X}$, and the {\em opposing} \talkingpoints{} $\fw{Y}$ belonging to 
its opponent $Y$. 
These are examples of \talkingpoints for the 
``Millennials'' debate:
\begin{center}
\begin{tabular}{| r  c |}
\hline
\small\textbf{Side} & \small\textbf{\Talkingpoints}\\
\hline
\small For & \small debt, boomer, college, reality\\%
\small Against & \small economy, volunteer, home, engage\\%
\hline
\end{tabular}
\end{center}
\xhdr{The flow of \talkingpoints{}}
A side can either promote its own \talkingpoints{}, address its opponent's points, or steer away from these initially salient ideas altogether.
We quantify the use of these strategies by comparing the airtime debaters devote to \talkingpoints{}. For a side $X$, 
let the {\em \selfcoverage} $\freqtp{r}{X}{X}$ be the fraction of content words uttered by $X$ in round $r$ that are among their own \talkingpoints{} $\fw{X}$; and the {\em \opponentcoverage} $\freqtp{r}{X}{Y}$ be the fraction of its content words covering opposing \talkingpoints{} $\fw{Y}$. %
\begin{figure}
\centering
\includegraphics[scale=.35]{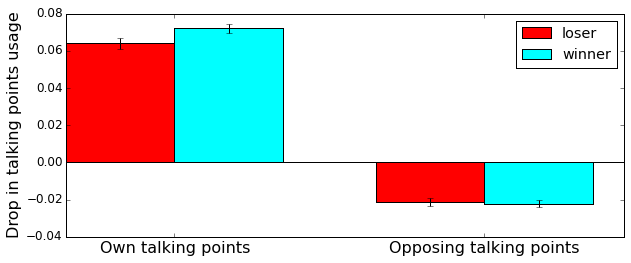}
\caption{The start of the debate's interactive stage triggers a drop in \selfcoverage ($>0$, indicated by leftmost two bars) and a rise in \opponentcoverage ($<0$, indicated by rightmost bars), with eventual winners showing a more pronounced drop in self-coverage (comparing the two bars on the left).}
\label{fig:talking_point_diffs}
\end{figure}

 Not surprisingly, we find that self-coverage dominates 
  during the discussion ($\freqtp{\Disc}{X}{X} > \freqtp{\Disc}{X}{Y}$, $p<0.001$). However, this does not mean debaters are simply giving monologues and ignoring each other: the effect of the interaction is reflected in a sharp drop in \selfcoverage and a rise in \opponentcoverage once the discussion round begins. Respectively, 
 $\freqtp{\Disc}{X}{X}<\freqtp{\Intro}{X}{X}$ and $\freqtp{\Disc}{X}{Y}>\freqtp{\Intro}{X}{Y}$, both $p<0.001$. Examples of self- and opponent-coverage of two \talkingpoint{}s in the ``Millennials'' debate from the introduction and discussion are given in Table \ref{tab:tp_example}.

Does the change in focus translate to any strategic advantages?  Figure \ref{fig:talking_point_diffs} suggests this is the case: the drop in self-coverage is slightly larger for the side that eventually wins the debate ($p = 0.08$).  
The drop in the sum of self- and \opponentcoverage is also larger for winning teams, suggesting that they are more likely to steer away from discussing \emph{any} talking points from either side ($p = 0.05$).

\xhdr{Identifying discussion points}
Having seen that debaters can benefit by shifting away from \talkingpoints that were salient during the introduction, we now examine the ideas that spontaneously arise to serve the discussion.  We model such {\em \discusspoints} as words introduced to the debate during the discussion by a debater and adopted by his opponents at least twice.\footnote{Ignoring single repetitions discards simple echoing of words used by the previous speaker.} This allows us to focus on words that become relevant to the conversation; only 3\% of all newly introduced words qualify, amounting to about 10 \discusspoints per debate.

\xhdr{The flow of \discusspoints{}}
The adoption of \discusspoints{} plays 
an important role in persuading the audience: during the discussion, eventual winners adopt more \discusspoints{} introduced by their opponents than eventual losers ($p < 0.01$). Two 
possible
 strategic interpretations emerge. From a topic control angle \cite{nguyen2014modeling}, perhaps losers are more successful at imposing their \discusspoints{} to gain control of the discussion. This view appears counterintuitive given work linking topic control to influence in other settings \cite{planalp1980not,rienks2006detection}. 

An alternative interpretation could be that
winners are more active than losers in contesting their opponents' points, a strategy that might play out favorably to the audience. A post-hoc manual examination supports this interpretation: 78\% of the valid \discusspoints 
 are picked up by the opposing side 
 in order to be 
 challenged;\footnote{Three annotators 
 (including one author) 
 informally annotated a random sample of 50 \discusspoints in the context of all 
 dialogue
  excerpts where the point was used. According to a majority vote, in 26 cases the opponents challenged the point, in 7 cases the point was supported, 4 cases were unclear, and in 13 cases the annotators deemed the \discusspoint invalid. We discuss 
 the last category
   in Section \ref{sec:discussion}.} 
 this strategy 
 is exemplified in Table \ref{tab:attack_example}. 
Overall, these observations tying the flow of discussion points to the debate's outcome
  suggest that winners are more successful at using the interaction to engage with their opponents' ideas. 
\begin{table}
\begin{center}
\begin{tabular}{| p{7cm} |}
\hline
\footnotesize
AGAINST: I would say [millennials] are effectively moving towards goals [...] it might seem like immaturity if you don't actually talk to millennials and look at the \textbf{statistics}. \\
\footnotesize
FOR: --actually, the numbers are showing [...] that it's worsening [...] Same \textbf{statistics}, dreadful \textbf{statistics}. \\
\hline
\footnotesize
AGAINST: [...] there's a incredible [sic] advantage that millennials have when it comes to social media [...] because we have an understanding of that landscape as \textbf{digital} natives [...] \\
\footnotesize
FOR: Generation X [...] is also known as the \textbf{digital} generation. The companies [...] that make you \textbf{digital} natives were all founded by [...] people in generation X. It's simply inaccurate every time somebody says that the millennial generation is the only generation [...]\\
\hline
\end{tabular}
\end{center}
\caption{Example \discusspoints{} introduced by the Against side in the ``Millennials'' debate. For each point, the first excerpt is the context in which the point was first mentioned by the Against side in the discussion, and the second excerpt shows the For side challenging the point later on. \label{tab:attack_example}}
\end{table}

\section{Predictive Power}
\label{sec:eval}
We evaluate the predictive power of our flow features in a binary classification setting: predict whether the For or Against side wins the debate.\footnote{The task is balanced: after removing three debates ending in a tie, we have 52 debates won by For and 53 by Against.}
This is a challenging task even for humans, thus the dramatic reveal at the end of each IQ2 debate that partly explains the popularity of the show.  Our goal here is limited to understanding which of the flow features that we developed carry predictive power.

\xhdr{Conversation flow features} We use all conversational features discussed above. For each side $X$ we include $\freqtp{\Disc}{X}{X}$, $\freqtp{\Disc}{X}{Y}$, and their sum. We also use the drop in \selfcoverage{} given by subtracting corresponding values for $\freqtp{\Intro}{\cdot}{\cdot}$, and the number of \discusspoints{} adopted by each side.  We call these the \textit{Flow} features.

\xhdr{Baseline features} To discard the possibility that our results are simply explained by debater verbosity, we use the number of words uttered and number of turns taken by each side (\emph{length}) as baselines. We also compare to a unigram baseline (\emph{BOW}).

\xhdr{Audience features} We use the counts of applause and laughter received by each side (described in Section \ref{sec:data}) as rough indicators of how well the audience can foresee a debate's outcome. 
Prediction accuracy is evaluated using a leave-one-out (LOO) approach. We use logistic regression; model parameters for each LOO train-test split are selected via 3-fold cross-validation on the training set. To find particularly predictive flow features, we also try using univariate feature selection on the flow features before the model is fitted in each split; we refer to this setting as \emph{Flow*}.\footnote{We optimize the regularizer ($\ell_1$ or $\ell_2$), and the value of the regularization parameter $C$ (between $10^{-5}$ and $10^5$). For Flow* we also optimize the number of features selected.} 

We find that conversation flow features obtain the best accuracy among all listed feature types (Flow: 63\%; Flow*: 65\%), performing significantly higher than a 50\% random baseline (binomial test $p < 0.05$), and comparable to audience features (60\%). In contrast, the length and BOW baselines do not perform better than chance. We note that \emph{Flow} features perform competitively despite being the only ones that do not factor in the concluding round.

The features selected most often in the Flow* task are: the number of \discusspoints{} adopted (with positive regression coefficients), the recall of \talkingpoints{} during the discussion round (negative coefficients), and the drop in usage of own \talkingpoints{} from introduction to discussion (positive coefficients). %
The relative importance of these features, which focus on the interaction between teams, suggests that audiences tend to favor debating strategies which emphasize the discussion.

\section{Further Related Work}
\label{sec:related}
Previous work on 
conversational structure has proposed
 approaches to model dialogue acts \cite{Samuel:1998,ritter2010unsupervised,Ferschke:2012} or disentangle interleaved conversations \cite{elsner2010disentangling,elsner2011disentangling}.
  Other research has considered the problem of 
  detecting
   conversation-level traits such as the presence of disagreements  \cite{allendetecting,wang2014piece} or the likelihood of relation dissolution \cite{niculae15betrayal}. At the participant level, several studies present approaches to identify ideological stances \cite{somasundaran2010recognizing,rosenthal2015couldn}, 
   using features 
	based on 
    participant interactions \cite{thomas2006get,sridhar2015joint},
	or extracting words and reasons characterizing a stance \cite{monroe2008fightin,nguyen2010analysis,hasan2014you}.
     In our setting, both the stances and the turn structure of a debate are known, allowing us to instead focus on the debate's outcome.

Existing research on argumentation 
strategies
 has largely focused on 
exploiting
 the structure of monologic arguments \cite{mochales2011argumentation}, like those of persuasive essays  \cite{feng2011classifying,stab2014identifying}.  In addition, \newcite{tan+etal:16a} has examined the effectiveness of arguments in the context of a forum where people invite others to challenge their opinions.%
 We complement this line of work by looking at the relative persuasiveness of participants in 
extended conversations as they exchange arguments over multiple turns.

Previous studies of influence in extended conversations have largely dealt with the political domain, examining moderated but relatively unstructured settings such as talk shows or presidential debates, and suggesting features like topic control \cite{nguyen2014modeling}, linguistic style matching \cite{romero2015mimicry} and turn-taking \cite{prabhakaran2013hand}.  With persuasion in mind, our work extends these studies to explore a new dynamic, the flow of ideas between speakers, in a highly structured setting that controls for confounding factors.

\section{Limitations and Future Work}
\label{sec:discussion}
This study opens several avenues for future research. One could explore more complex representations of \talkingpoints{} and \discusspoints{}, for instance using topic models or word embeddings. 
Furthermore, augmenting
 the flow of content in a conversation 
 with
  the speakers' linguistic 
  choices
   could 
  better capture their intentions.
    In addition, it would be interesting to study the interplay between our conversational flow features and relatively monologic features that consider the argumentative and rhetorical traits of each side separately. More explicitly comparing and contrasting monologic and interactive dynamics could 
    lead to
    better models of conversations. Such approaches could also help clarify some of the intuitions about conversations explored in this work, particularly that engaging in dialogue carries different strategic implications from self-promotion. 

Our focus in this paper is on capturing and understanding conversational flow. We hence make some simplifying assumptions that could be refined in future work. For instance, by using a basic unigram-based definition of \discusspoints{}, we do not account for the context or semantic sense in which these points occur. In particular, our annotators found that a significant proportion of the \discusspoints{} under our definition actually referred to differing ideas in the various contexts in which they appeared. We expect that improving our retrieval model will also improve the robustness of our idea flow analysis. A better model of \discusspoints{} could also provide more insight into the role of these points in persuading the audience. 

While Oxford-style debates are a particularly convenient setting for 
studying the effects of conversational flow,
 our dataset is limited in terms of size. It would be worthwhile to examine the flow features we developed in the context of settings with richer incentives beyond persuading an audience, such as in the semi-cooperative environment of Wikipedia talk pages. Finally, our methodology could point to applications in areas such as education and cooperative work, where it is key to establish the link between conversation features and an interlocutor's ability to convey their point \cite{niculae16}.  

\xhdr{Acknowledgements}  We 
thank the reviewers and
 V. Niculae for their helpful comments, and I. Arawjo and D. Sedra for annotations. This work was supported in part by a Google Faculty Research Award.

\bibliography{refs-iq2}
\bibliographystyle{naaclhlt2016}

\end{document}